%% file: paper3421.tex
\documentclass[runningheads]{llncs}
\usepackage[T1]{fontenc}
\usepackage{graphicx}
\usepackage{multirow}
\usepackage{booktabs}
\usepackage{amsmath,amssymb,amsfonts}
\usepackage[table,xcdraw]{xcolor}
\usepackage[colorlinks]{hyperref}
\usepackage{color, colortbl}
\usepackage{footmisc}
\usepackage{subfig}
\usepackage{float}

\begin{document}

\input{0_title_page}
\input{1_intro}
\input{2_related_work}

\input{3_method}

\input{4_experiments}
\input{5_results}

\input{6_conclusion}
\input{7_acknowledgements}

\bibliographystyle{splncs04}
\bibliography{bibliography3421}
\appendix
\input{8_appendix}

\end{document}

%% file: 0_title_page.tex
\title{vox2vec: A Framework for Self-supervised Contrastive Learning of Voxel-level Representations in Medical Images}

\titlerunning{vox2vec}

\author{
    Mikhail Goncharov \inst{1} \and
    Vera Soboleva \inst{2} \and
    Anvar Kurmukov \inst{3} \and 
    Maxim Pisov \inst{4} \and
    Mikhail Belyaev \inst{1,3}
}

\authorrunning{M. Goncharov et al.}

\institute{
    Skolkovo Institute of Science and Technology, Moscow, Russia
    \and
    Artificial Intelligence Research Institute (AIRI), Moscow, Russia
    \and
    Institute for Information Transmission Problems, Moscow, Russia
    \and
    IRA-Labs, Moscow, Russia
    \\
    \email{Mikhail.Goncharov2@skoltech.ru}
}

\maketitle

\begin{abstract}

This paper introduces \texttt{vox2vec} --- a contrastive method for self-supervised learning (SSL) of voxel-level representations. \texttt{vox2vec} representations are modeled by a Feature Pyramid Network (FPN): a voxel representation is a concatenation of the corresponding feature vectors from different pyramid levels. The FPN is pre-trained to produce similar representations for the same voxel in different augmented contexts and distinctive representations for different voxels. This results in unified multi-scale representations that capture both global semantics (e.g., body part) and local semantics (e.g., different small organs or healthy versus tumor tissue). We use \texttt{vox2vec} to pre-train a FPN on more than 6500 publicly available computed tomography images. We evaluate the pre-trained representations by attaching simple heads on top of them and training the resulting models for 22 segmentation tasks. We show that \texttt{vox2vec} outperforms existing medical imaging SSL techniques in three evaluation setups: linear and non-linear probing and end-to-end fine-tuning. Moreover, a non-linear head trained on top of the frozen \texttt{vox2vec} representations achieves competitive performance with the FPN trained from scratch while having $50$ times fewer trainable parameters. The code is available at \url{https://github.com/mishgon/vox2vec}.

\keywords{
    Contrastive Self-Supervised Representation Learning \and
    Medical Image Segmentation
}

\end{abstract}

%% file: 1_intro.tex
\section{Introduction}

Medical image segmentation often relies on supervised model training \cite{nnunet}, but this approach has limitations. Firstly, it requires costly manual annotations. Secondly, the resulting models may not generalize well to unseen data domains. Even small changes in the task may result in a significant drop in performance, requiring re-training from scratch \cite{kondrateva2022neglectable}.

Self-supervised learning (SSL) is a promising solution to these limitations. SSL pre-trains a model backbone to extract informative representations from unlabeled data. Then, a simple linear or non-linear head on top of the frozen pre-trained backbone can be trained for various downstream tasks in a supervised manner (linear or non-linear probing). Alternatively, the backbone can be fine-tuned for a downstream task along with the head. Pre-training the backbone in a self-supervised manner enables scaling to larger datasets across multiple data and task domains. In medical imaging, this is particularly useful given the growing number of available datasets.

In this work, we focus on contrastive learning \cite{contrastive,simclr}, one of the most effective approaches to SSL in computer vision. 
In contrastive learning, the model is trained to produce similar vector representations for augmented views of the same image and dissimilar representations for different images. Contrastive methods can also be used to learn dense, i.e., patch-level or even pixel- or voxel-level representations: pixels of augmented image views from the same region of the original image should have similar representations, while different pixels should have dissimilar ones \cite{vader}.

Several works have implemented contrastive learning of dense representations in medical imaging \cite{3d_ssl,global_and_local,transvw,swin_unetr,sam}. Representations in \cite{3d_ssl,global_and_local} do not resolve nearby voxels due to the negative sampling strategy and the architectural reasons. This makes them unsuitable for full-resolution segmentation, especially in linear and non-linear probing regimes. In the current SotA dense SSL methods \cite{transvw,swin_unetr}, authors employ restorative learning in addition to patch-level contrastive learning, in order to pre-train voxel-level representations in full-resolution. In \cite{sam}, separate global and voxel-wise representations are learned in a contrastive manner to implement efficient dense image retrieval. 

The common weakness of all the above works is that they do not evaluate their SSL models in linear or non-linear probing setups, even though these setups are de-facto standards for evaluation of SSL methods in natural images \cite{simclr,mae,vader}. Moreover, fine-tuned models can deviate drastically from their pre-trained states due to catastrophical forgetting \cite{catastrophic}, while models trained in linear or non-linear probing regimes are more robust as they have several orders of magnitude fewer trainable parameters.

Our contributions are threefold. \textbf{First}, we propose \texttt{vox2vec}, a framework for contrastive learning of voxel-level representations. Our simple negative sampling strategy and the idea of storing voxel-level representations in a feature pyramid form result in high-dimensional, fine-grained, multi-scale representations suitable for the segmentation of different organs and tumors in full resolution. \textbf{Second}, we employ \texttt{vox2vec} to pre-train a FPN architecture on a diverse collection of six unannotated datasets, totaling over 6,500 CT images of the thorax and abdomen. We make the pre-trained model publicly available to simplify the reproduction of our results and to encourage practitioners to utilize this model as a starting point for the segmentation algorithms training. \textbf{Finally}, we compare the pre-trained model with the baselines on 22 segmentation tasks on seven CT datasets in three setups: linear probing, non-linear probing, and fine-tuning. We show that \texttt{vox2vec} performs slightly better than SotA models in the fine-tuning setup and outperforms them by a huge margin in the linear and non-linear probing setups. To the best of our knowledge, this is the first successful attempt to evaluate dense SSL methods in the medical imaging domain in linear and non-linear probing regimes.

%% file: 2_related_work.tex
\section{Related work}

In recent years, self-supervised learning in computer vision has evolved from simple pretext tasks like Jigsaw Puzzles \cite{jigsaw}, Rotation Prediction \cite{rotation}, and Patch Position Prediction \cite{context_patch} to the current SotA methods such as restorative autoencoders \cite{mae} and contrastive \cite{simclr} or non-contrastive \cite{simsiam} joint embedding methods.

Several methods produce dense or pixel-wise vector representations~\cite{vader,propagate_yourself,vicregl} to pre-train models for downstream tasks like segmentation or object detection. In \cite{vader}, pixel-wise representations are learned by forcing local features to remain constant over different viewing conditions. This means that matching regions describing the same location of the scene on different views should be positive pairs, while non-matching regions should be negative pairs. In \cite{propagate_yourself}, authors define positive and negative pairs as spatially close and distant pixels, respectively. While in \cite{vicregl}, authors minimize the mean square distance between matched pixel embeddings, simultaneously preserving the embedding variance along the batch and decorrelating different embedding vector components.

The methods initially proposed for natural images are often used to pre-train models on medical images. In \cite{3d_ssl}, authors propose the 3D adaptation of Jigsaw Puzzle, Rotation Prediction, Patch Position Prediction, and image-level contrastive learning. Another common way for pre-training on medical images is to combine different approaches such as rotation prediction \cite{swin_unetr}, restorative autoencoders \cite{transvw,swin_unetr}, and image-level contrastive learning \cite{transvw,swin_unetr}.

Several methods allows to obtain voxel-wise features. The model \cite{sam} maximizes the consistency of local features in the intersection between two differently augmented images. The algorithm \cite{sam} was mainly proposed for image retrieval and uses only feature representations in the largest and smallest scales in separate contrastive losses, while \texttt{vox2vec} produce voxels' representations via concatenation of feature vectors from a feature pyramid and pre-train them in a unified manner using a single contrastive loss. Finally, a number of works propose semi-supervised contrastive learning methods \cite{semi1}, however, they require additional task-specific manual labeling.

%% file: 3_method.tex
\section{Method}
\label{seq:method}

\begin{figure}[t]
\begin{center}
    \includegraphics[width=1\linewidth]{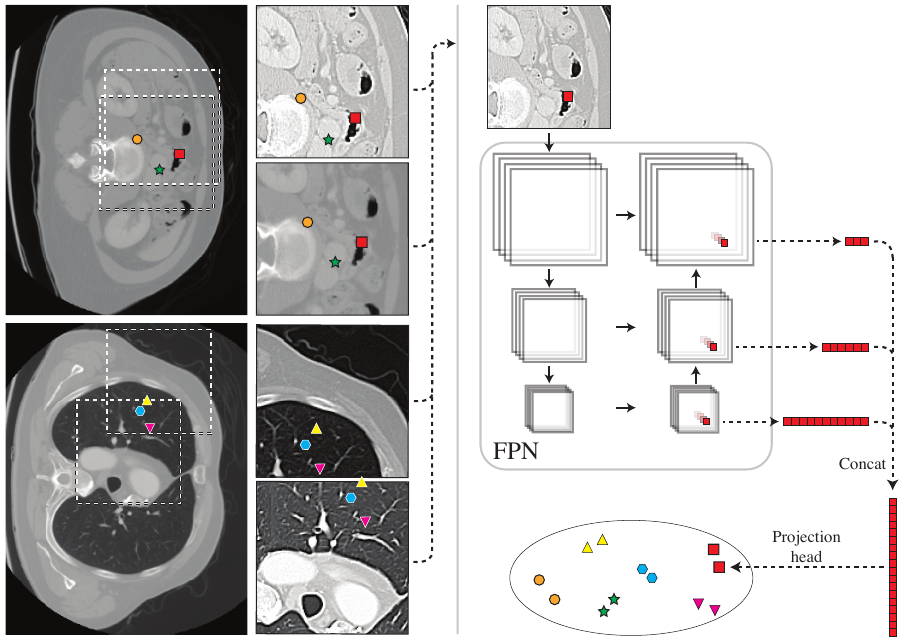}
    \caption{Illustration of the \texttt{vox2vec} pre-training pipeline. Left: two overlapping augmented 3D patches are sampled from each volume in a batch. Markers of the same color and shape denote positive pairs of voxels. Right: voxel-level representations are obtained via the concatenation of corresponding feature vectors from different levels of the FPN. Finally, the representations are projected to the space where contrastive loss is computed.}
    \label{fig:method}
\end{center}
\end{figure}

In a nutshell, \texttt{vox2vec} pre-trains a neural network to produce similar representations for the same voxel placed in different contexts (positive pairs) and predict distinctive representations for different voxels (negative pairs). In the following Sections~\ref{subseq:method:sampling},~\ref{subseq:method:architecture},~\ref{subseq:method:loss}, we describe in detail the main components of our method: 1) definition and sampling of positive and negative pairs of voxels; 2) modeling voxel-level representations via a neural network; 3) computation of the contrastive loss. The whole pre-training pipeline is schematically illustrated in Figure~\ref{fig:method}. We also describe the methodology of the evaluation of the pre-trained representations on downstream segmentation tasks in Section~\ref{subseq:method:evaluation}.

\subsection{Sampling of Positive and Negative Pairs}
\label{subseq:method:sampling}

We define a \textit{positive pair} as any pair of voxels that correspond to the same location in a given volume. Conversely, we call a \textit{negative pair} any pair of voxels that correspond to different locations in the same volume as well as voxels belonging to different volumes.

Figure~\ref{fig:method} (left) illustrates our strategy for positive and negative pairs sampling. For a given volume, we sample two overlapping 3D patches of size $(H, W, D)$.
We apply color augmentations to them, including random gaussian blur, random gaussian sharpening, adding random gaussian noise, clipping the intensities to the random Hounsfield window, and rescaling them to the $(0, 1)$ interval. 
Next, we sample $m$ different positions from the patches' overlapping region. Each position yields a pair of voxels --- one from each patch, which results in a total of $m$ positive pairs of voxels. At each pre-training iteration, we repeat this procedure for $n$ different volumes, resulting in $2 \cdot n$ patches containing $N = n \cdot m$ positive pairs.
Thus, each sampled voxel has one positive counterpart and forms negative pairs with all the remaining $2N - 2$ voxels.

In our experiments we set $(H, W, D) = (128, 128, 32)$, $n = 10$ and $m = 1000$.

We exclude the background voxels from the sampling and do not penalize their representations. We obtain the background voxels by using a simple two-step algorithm: 1) thresholding voxels with an intensity less than $-500$ HU; 2) keep voxels from the same connected component as the corner voxel of the CT volume, using a flood fill algorithm.

\subsection{Architecture}
\label{subseq:method:architecture}
A standard architecture for voxel-wise prediction is 3D UNet \cite{unet}. UNet's backbone returns a feature map of the same resolution as the input patch. However, our experiments show that this feature map alone is insufficient for modeling self-supervised voxel-level representations. The reason is that producing a feature map with more than $100$ channels in full resolution is infeasible due to memory constraints. Meanwhile, to be suitable for many downstream tasks, representations should have a dimensionality of about $1000$, as in \cite{simclr}.

To address this issue, we utilize a 3D FPN architecture instead of a standard 3D UNet. FPN returns voxel-level representations in the form of a feature pyramid. The pyramid's base is a feature map with $16$ channels of the same resolution as the input patch. Each next pyramid level has twice as many channels and two times lower resolution than the previous one. Each voxel's representation is a concatenation of the corresponding feature vectors from all the pyramid levels. We use FPN with six pyramid levels, which results in $1008$-dimensional representations. See Figure~\ref{fig:method} (right) for an illustration.

\subsection{Loss Function}
\label{subseq:method:loss}

At each pre-training iteration, we fed $2 \cdot n$ patches to the FPN and obtain the representations for $N$ positive pairs of voxels. We denote the representations in $i$-th positive pair as $h_i^{(1)}$ and $h_i^{(2)}$, $i = 1, \ldots, N$. Following \cite{simclr}, instead of penalizing the representations directly, we project them on $128$-dimensional unit sphere via a trainable $3$-layer perceptron $g(\cdot)$ followed by l$2$-normalization: $z_i^{(1)} = g(h_i^{(1)}) / \|g(h_i^{(1)})\|$, $z_i^{(2)} = g(h_i^{(2)}) / \|g(h_i^{(2)})\|$, $i = 1, \ldots, N$. Similar to \cite{simclr} we use the InfoNCE loss as a contrastive objective: $\mathcal{L} = \sum_{i = 1}^{N} \sum_{k \in \{1, 2\}}\mathcal{L}_i^{k}$, where
$$
\mathcal{L}_i^{k} = -\log\frac{\exp(\langle z_i^{(1)}, z_i^{(2)} \rangle / \tau)}{\exp(\langle z_i^{(1)}, z_i^{(2)} \rangle / \tau) + \sum_{j \in \{1, \ldots, N\} \setminus \{i\}} \sum_{l \in \{1, 2\}} \exp(\langle z_i^{(k)}, z_j^{(l)} \rangle / \tau)}.
$$

\subsection{Evaluation protocol}
\label{subseq:method:evaluation}
We evaluate the quality of self-supervised voxel-level representations on downstream segmentation tasks in three setups: 1) linear probing, 2) non-linear probing, and 3) end-to-end fine-tuning.

Linear or non-linear probing means training a voxel-wise linear or non-linear classifier on top of the frozen representations. If the representations are modeled by the UNet model, such classifier can be implemented as one or several $1 \times 1$ convolutional layers with a kernel size $1$ on top of the output feature map.
A linear voxel-wise head (linear FPN head) can be implemented as follows. Each pyramid level is separately fed to its own convolutional layer with kernel size 1. Then, as the number of channels on all pyramid levels has decreased, they can be upsampled to the full resolution and summed up. This operation is equivalent to applying a linear classifier to FPN voxel-wise representations described in Section~\ref{subseq:method:architecture}. Linear FPN head has four orders of magnitude fewer parameters than FPN.
The architecture of the non-linear voxel-wise head replicates the UNet's decoder but sets the kernel size of all convolutions to $1$. It has 50 times fewer parameters than the entire FPN architecture.

In the end-to-end fine-tuning setup, we attach the voxel-wise non-linear head, but in contrast to the non-linear probing regime, we also train the backbone.

%% file: 4_experiments.tex
\section{Experiments}
\subsection{Pre-training}

We use \texttt{vox2vec} to pre-train both FPN and UNet models (further \texttt{vox2vec}-FPN and \texttt{vox2vec}-UNet) in order to ablate the effect of using a feature pyramid instead of single full-resolution feature map for modeling voxel-wise representations. For pre-training, we use $6$ public CT datasets~\cite{amos,flare,nlst,nsclc,lidc,midrc}, totaling more than $6550$ CTs, covering abdomen and thorax domains. We do not use the annotations for these datasets during the pre-training stage. Pre-processing includes the following steps: 1) cropping to the minimal volume containing all the voxels with the intensity greater than $-500$ HU; 2) interpolation to the voxel spacing of $1\times 1 \times 2$~mm$^3$ (intensities are clipped and rescaled at the augmentation step, see Section~\ref{subseq:method:sampling}). We pre-train both models for $100$K batches using the Adam optimizer \cite{adam} with a learning rate of $0.0003$. Both models are trained on a single A100-40Gb GPU for an average of $3$ days. Further details about the pre-training setup can be found in Supplementary materials.

\subsection{Evaluation}

We evaluate our method on the Beyond the Cranial Vault Abdomen (BTCV)~\cite{btcv} and Medical Segmentation Decathlon (MSD)~\cite{msd} datasets. The BTCV dataset consists of $30$ CT scans along with $13$ different organ annotations. We test our method on $6$ CT MSD datasets, which include $9$ different organ and tumor segmentation tasks. A $5$ fold cross-validation is used for BTCV experiments, and a $3$ fold cross-validation for MSD experiments. The segmentation performance of each model on BTCV and MSD datasets is evaluated by the Dice score.

For our method, the pre-processing steps are the same for all datasets, as at the pre-training stage, but in addition, intensities are clipped to $(-1350, 1000)$ HU window and rescaled to $(0, 1)$.

We compare our results with the current state-of-the-art self-supervised methods~\cite{swin_unetr,transvw} in medical imaging. The pre-trained weights for the SwinUNETR encoder and TransVW UNet are taken from the official repositories of corresponding papers. In these experiments, we keep the crucial pipeline hyperparameters (e.g., spacing, clipping window, patch size) the same as in the original works. To evaluate the pre-trained SwinUNETR and TransVW in linear and non-linear probing setups, we use similar linear and non-linear head architectures as for \texttt{vox2vec}-FPN (see Section~\ref{subseq:method:evaluation}). SwinUNETR and TransVW cost $391$ GFLOPs and $1.2$ TFLOPS, correspondingly, compared to $115$ GFLOPs of \texttt{vox2vec}-FPN.

We train all models for $45000$ batches of size $7$ (batch size for SwinUNETR is set to $3$ due to memory constraints), using the Adam optimizer with a learning rate of $0.0003$. In the fine-tuning setup, we freeze the backbone for the first $15000$ batches and then exponentially increase the learning rate for the backbone parameters from $0.00003$ up to $0.0003$ during $1200$ batches. 

%% file: 5_results.tex
\section{Results}
\renewcommand{\arraystretch}{1.33}

\begin{figure}[]
  \centering
  \begin{minipage}[valign=t]{0.51\textwidth}
   \centering
    \captionof{table}{Average cross validation Dice scores on BTCV multi-organ segmentation dataset.}
    \label{tab:btcv}
    \resizebox{\textwidth}{!}{\begin{tabular}{ccccccccccccc}
    \toprule
    model & Sp   & Kid  & Gb   & Es   & Li   & St   & Aor  & IVC  & PSV  & Pa & AG & Avg  \\
    \midrule
    \multicolumn{13}{c}{\cellcolor[HTML]{EFEFEF}\textbf{from scratch}} \\
    TransVW UNet & 79.2 & 82.7 & 43.9 & 65.9 & 83.7 & 62.1 & 86.6 & 76.9 & 61.3 & 56.7 & 51.4 & 68.0 $\pm$ 2.1 \\
    SwinUNETR    & 90.8 & 87.8 & 60.4 & 69.8 & 94.7 & 79.8 & 88.0 & 81.8 & 67.7 & 69.6 & 61.5 & 77.0 $\pm$ 2.5\\
    UNet         & 91.1 & 88.5 & 58.8 & 72.3 & 96.0 & \textbf{83.8} & 89.0 & 83.2 & 68.3 & 70.4 & 63.2 & 78.2 $\pm$ 2.3 \\
    FPN         & \textbf{92.4} & 89.5 & \textbf{60.9} & 70.1 & \textbf{96.3} & 82.7 & 90.1 & \textbf{83.9} & 69.0 & 71.8 & 62.5 & 78.5 $\pm$ 2.2 \\
    \multicolumn{13}{c}{\cellcolor[HTML]{EFEFEF}\textbf{linear probing}} \\
    TransVW & 34.4 & 25.7 & 8.9 & 34.4 & 56.8 & 12.1 & 47.2 & 19.0 & 18.8 & 8.2 & 20.6 & 25.6 $\pm$ 1.1 \\
    SwinUNETR    & 44.4 & 38.3 & 7.6 & 23.7 & 72.4 & 17.8 & 36.6 & 26.9 & 19.4 & 3.6 & 11.8 & 27.1 $\pm$ 2.4 \\
    random-FPN    & 68.0 & 61.2 & 30.0 & 38.0 & 81.6 & 45.3 & 65.0 & 52.4 & 27.7 & 22.9 & 26.0 & 46.6 $\pm$ 3.0 \\
    vox2vec-UNet & 79.4 & 79.8 & 29.9 & 37.7 & 90.5 & 62.5 & 78.8 & 70.8 & 36.0 & 40.9 & 33.6 & 57.9 $\pm$ 2.0\\
    vox2vec-FPN  & 83.7 & 84.0 & 43.7 & 58.0 & 93.1 & 67.5 & 85.6 & 77.5 & 56.6 & 58.8 & 53.3 & 69.2 $\pm$ 1.2 \\
    \multicolumn{13}{c}{\cellcolor[HTML]{EFEFEF}\textbf{non-linear probing}} \\
    TransVW      & 24.9 & 31.5 & 6.7 & 28.1 & 45.1 & 9.0 & 44.9 & 27.2 & 19.0 & 7.2 & 15.4 & 23.5 $\pm$ 2.7 \\
    random-FPN   & 76.7 & 67.0 & 34.1 & 47.1 & 83.7 & 52.8 & 70.2 & 57.5 & 30.2 & 28.6 & 31.5 & 52.1 $\pm$ 4.9 \\
    SwinUNETR    & 77.0 & 74.4 & 48.1 & 52.1 & 87.0 & 53.7 & 73.5 & 58.1 & 47.2 & 35.3 & 39.9 & 58.5 $\pm$ 2.6 \\
    vox2vec-UNet & 80.3 & 81.4 & 34.1 & 42.7 & 91.1 & 64.0 & 79.6 & 71.6 & 42.7 & 43.3 & 37.6 & 60.6 $\pm$ 3.0 \\
    vox2vec-FPN  & 91.0 & 89.2 & 50.7 & 67.5 & 95.3 & 78.2 & 89.4 & 80.7 & 64.9 & 66.1 & 59.9 & 75.5 $\pm$ 1.7 \\
    \multicolumn{13}{c}{\cellcolor[HTML]{EFEFEF}\textbf{fine-tuning}} \\
    TransVW      & 77.8 & 80.7 & 42.9 & 66.5 & 83.6 & 59.3 & 86.2 & 77.3 & 63.7 & 54.4 & 54.0 & 67.8 $\pm$ 1.9 \\
    SwinUNETR    & 84.2 & 86.7 & 58.4 & 70.4 & 94.5 & 76.0 & 87.7 & 82.1 & 67.0 & 69.8 & 61.0 & 75.8 $\pm$ 3.3 \\
    vox2vec-UNet & 91.4 & 90.1 & 52.3 & 72.5 & 95.8 & 83.0 & 89.9 & 82.6 & 66.5 & 71.1 & 61.8 & 77.6 $\pm$ 1.0\\
    vox2vec-FPN  & 91.4 & \textbf{90.7} & 59.5 & \textbf{72.7} & \textbf{96.3} & 83.2 & \textbf{91.3} & \textbf{83.9} & \textbf{69.2} & \textbf{73.9} & \textbf{65.2} & \textbf{79.5 $\pm$ 1.3} \\ 
    \bottomrule
    \end{tabular}}
  \end{minipage}
  \hfill
  \begin{minipage}[valign=t]{0.47\textwidth}
    \includegraphics[width=\textwidth]{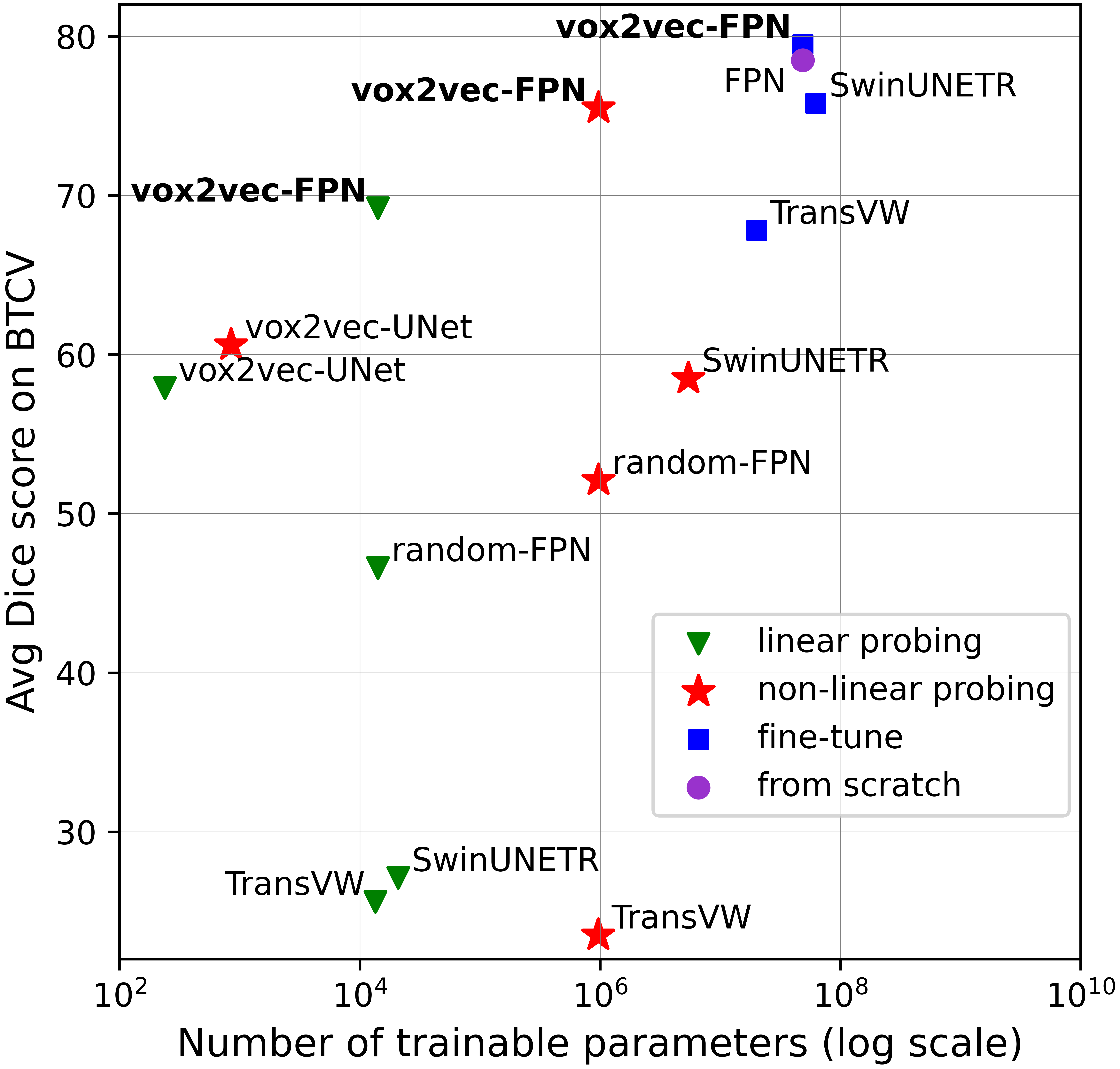}
    \caption{Dice score on BTCV cross-validation averaged for all organs w.r.t. the number of trainable paramaters of different models in different evaluation setups.}
    \label{fig:params}
  \end{minipage}
\end{figure}

The mean value and standard deviation of Dice score across $5$ folds on the BTCV dataset for all models in all evaluation setups are presented in Table~\ref{tab:btcv}. \texttt{vox2vec}-FPN performs slightly better than other models in the fine-tuning setup. However, considering the standard deviation, all the fine-tuned models perform on par with their counterparts trained from scratch.

Nevertheless, \texttt{vox2vec}-FPN significantly outperforms other models in linear and non-linear regimes. On top of that, we observe that in non-linear probing regime, it performs (within the standard deviation) as well as the FPN trained from scratch while having $\text{x}50$ times fewer trainable parameters (see Figure~\ref{fig:params}). We demonstrate an example of the excellent performance of \texttt{vox2vec}-FPN in both linear and non-linear probing regimes in Supplementary materials.

We reproduce the key results on MSD challenge CT datasets, which contain tumor and organ segmentation tasks. Table~\ref{tab:msd} shows that in the \texttt{vox2vec} representation space, organ voxels can be separated from tumor voxels with a quality comparable to the model trained from scratch. A t-SNE embedding of vox2vec representations on MSD is available in the Supplementary materials.

\begin{table}[h!]
\centering
\caption{Cross validation Dice score on CT tasks of MSD challenge.}\label{tab:msd}
\resizebox{0.6\textwidth}{!}{\begin{tabular}{cccccccccc}
\toprule
\multicolumn{1}{c|}{}   & \multicolumn{2}{c|}{Liver}   & \multicolumn{1}{c|}{Lung} & \multicolumn{2}{c|}{Pancreas} & \multicolumn{2}{c|}{Hepatic vessel} & \multicolumn{1}{c|}{Spleen} & Colon \\
\midrule
\multicolumn{1}{c|}{model} & organ & \multicolumn{1}{c|}{tumor} & 
\multicolumn{1}{c|}{tumor} & organ & \multicolumn{1}{c|}{tumor} & organ & \multicolumn{1}{c|}{tumor} & \multicolumn{1}{c|}{organ} & cancer \\
\multicolumn{10}{c}{\cellcolor[HTML]{EFEFEF}\textbf{from scratch}} \\
\multicolumn{1}{c|}{FPN} & 94.4 & \multicolumn{1}{c|}{44.6} & \multicolumn{1}{c|}{53.1} & \textbf{77.1} & \multicolumn{1}{c|}{28.0} & 53.7 & \multicolumn{1}{c|}{49.4} & \multicolumn{1}{c|}{96.0} & \textbf{32.2} \\
\multicolumn{10}{c}{\cellcolor[HTML]{EFEFEF}\textbf{non-linear probing}} \\
\multicolumn{1}{c|}{vox2vec-FPN} & 94.7 & \multicolumn{1}{c|}{43.9} & \multicolumn{1}{c|}{49.5} & 71.4 & \multicolumn{1}{c|}{28.5} & 58.1 & \multicolumn{1}{c|}{54.8} & \multicolumn{1}{c|}{95.1} & 24.8 \\
\multicolumn{10}{c}{\cellcolor[HTML]{EFEFEF}\textbf{fine-tuning}} \\
\multicolumn{1}{c|}{SwinUNETR} & 95.0 & \multicolumn{1}{c|}{49.3} & \multicolumn{1}{c|}{55.2} & 75.2 & \multicolumn{1}{c|}{\textbf{35.9}} & \textbf{60.9} & \multicolumn{1}{c|}{57.5} & \multicolumn{1}{c|}{95.5} & 29.2 \\
\multicolumn{1}{c|}{vox2vec-FPN} & \textbf{95.6} & \multicolumn{1}{c|}{\textbf{51.0}} & \multicolumn{1}{c|}{\textbf{56.6}} & 77.0 & \multicolumn{1}{c|}{31.8} & 59.5 & \multicolumn{1}{c|}{\textbf{62.4}} & \multicolumn{1}{c|}{\textbf{96.1}} & 30.1 \\
\bottomrule
\end{tabular}}
\end{table}

%% file: 6_conclusion.tex
\section{Conclusion}

In this work, we present \texttt{vox2vec} --- a self-supervised framework for voxel-wise representation learning in medical imaging. Our method expands the contrastive learning setup to the feature pyramid architecture allowing to pre-train effective representations in full resolution. By pre-training a FPN backbone to extract informative representations from unlabeled data, our method scales to large datasets across multiple task domains. We pre-train a FPN architecture on more than 6500 CT images and test it on various segmentation tasks, including different organs and tumors segmentation in three setups: linear probing, non-linear probing, and fine-tuning. Our model outperformed existing methods in all regimes. Moreover, \texttt{vox2vec} establishes a new state-of-the-art result on the linear and non-linear probing scenarios. 

Still, this work has a few limitations to consider. We plan to investigate further how the performance of \texttt{vox2vec} scales with the increasing size of the pre-training dataset and the pre-trained architecture size. Another interesting research direction is exploring the effectiveness of \texttt{vox2vec} in the domain adaptation and few-shot learning scenarios.

%% file: 7_acknowledgements.tex
\paragraph{Acknowledgements.}

This work was supported by the Russian Science Foundation grant number 20-71-10134.

%% file: 8_appendix.tex
\section{Pre-training setup}
\renewcommand{\arraystretch}{1.2}

\begin{table}[H]
\caption[]{Parameters of the augmentations used for positive pairs sampling, described in the main paper, Section~3.1. We use the MONAI notation for the augmentations and their parameters\footnotemark. Random gaussian blur and random gaussian sharpening are applied in the axial plane. With a probability of $0.5$, neither gaussian blur nor gaussian sharpening is used. Otherwise, one of them is equally likely to be applied. Clipping window is either fixed with a probability of $0.2$, or sampled from uniform distribution with a probability of $0.8$.}\label{tab:msd}
\begin{tabular}{ll}
\hline
\textbf{Augmentation}      & \textbf{Parameters}                               \\ \hline
RandGaussianSmooth     & $\sigma^{x,y,z} = (0.25, 1.5)$, $prob = 0.5$  \\
RandGaussianSharpen & $\sigma_1^{x,y,z} = (0.5, 1.0)$, $\sigma_2^{x,y,z} = 0.5$, $\alpha = (10.0, 30.0)$, $prob = 0.5$ \\
RandGaussianNoise      & $std \sim  \mathcal{U}(0, 30)$, $prob = 0.5$ \\
ScaleIntensityRanged   & $prob = 0.2: (a_{min}, a_{max}) = (-1350, 1000),$ \\ 
 & $prob = 0.8: (a_{min}, a_{max}) \sim (\mathcal{U}(-1350, -1000)$, $\mathcal{U}(300, 1000))$, \\ 
 & $(b_{min}, b_{max}) = (0, 1)$, $clip = True$ \\ \hline
\end{tabular}
\end{table}
\footnotetext{\url{https://docs.monai.io/en/stable/transforms.html\#vanilla-transforms}}

\renewcommand{\arraystretch}{1.0}
\begin{table}[]
    \centering
    \caption{Overview of the datasets used for the pre-training of the \texttt{vox2vec} models and for the evaluation of all the models.}\label{tab:data}
    \begin{tabular}{lccc}
    \toprule
         Dataset & Number of volumes & ROI & Original annotation  \\
         \midrule
         \multicolumn{4}{c}{\cellcolor[HTML]{EFEFEF}\textbf{self-supervised pre-training}}\\
         \multicolumn{1}{l}{AMOS} & 500 & Abdomen & 15 abdominal organs \\
         \multicolumn{1}{l}{FLARE2022} & 2000 & Abdomen & $-$ \\ 
         \multicolumn{1}{l}{NLST} & 2500 & Thorax & $-$ \\
         \multicolumn{1}{l}{MIDRC-RICORD} & 120 & Thorax & COVID-19\\
         \multicolumn{1}{l}{NSCLC-Radiomic} & 422 & Thorax & Lung cancer \\
         \multicolumn{1}{l}{LIDC} & 1018 & Thorax & Lung nodules\\
         \multicolumn{4}{c}{\cellcolor[HTML]{EFEFEF}\textbf{evaluation}} \\
         \multicolumn{1}{l}{BTCV} & 30 & Abdomen & 13 abdominal organs \\
         \multicolumn{1}{l}{MSD-Liver} & 131 & Abdomen & Liver, tumour \\
         \multicolumn{1}{l}{MSD-Lung} & 64 & Thorax & Lung, tumour \\
         \multicolumn{1}{l}{MSD-Pancreas} & 282 & Abdomen & Liver, tumour \\
         \multicolumn{1}{l}{MSD-Hepatic} & 303 & Abdomen & Hepatic Vessels, tumour \\
         \multicolumn{1}{l}{MSD-Spleen} & 41 & Abdomen & Spleen \\
         \multicolumn{1}{l}{MSD-Colon} & 126 & Abdomen & Colon cancer \\
         \bottomrule
    \end{tabular}
    \label{tab:my_label}
\end{table}

\section{Qualitative results}

\begin{figure}[H]
    \centering
    \includegraphics[width=0.9\textwidth]{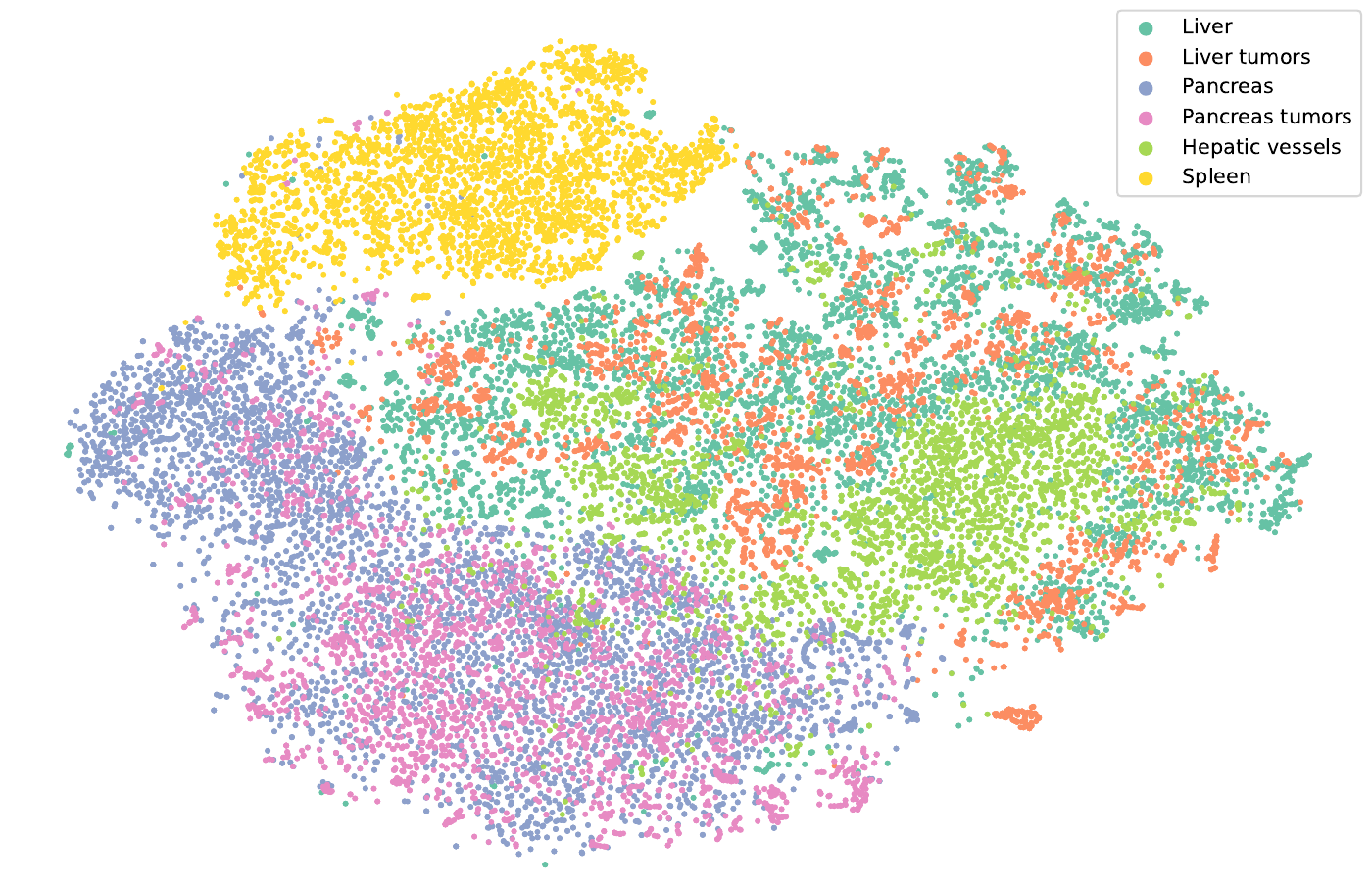}
    \caption{t-SNE plot of the \texttt{vox2vec} representations of voxels belonging to the different semantic classes (organs and tumors). For each class, more than 5000 voxels belonging to the corresponding segmentation masks were randomly sampled from different volumes from MSD datasets.}
    \label{fig:tsne}
\end{figure}

\begin{figure}[H]
    \centering
    \includegraphics[width=\textwidth]{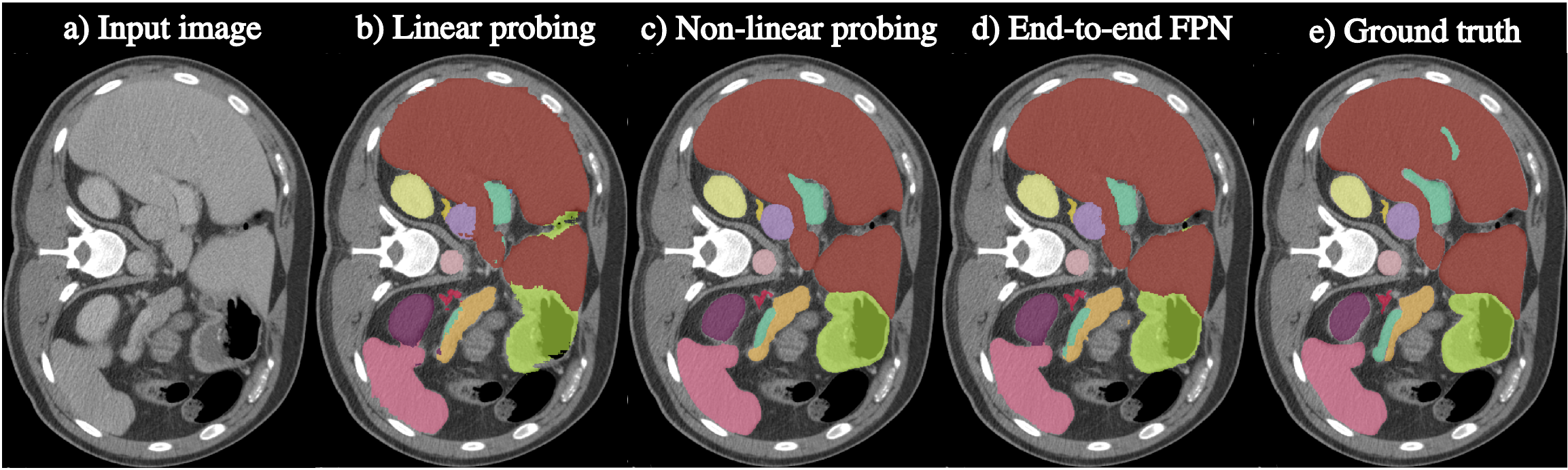}
    \caption{An example from a test fold of the BTCV dataset, showing that linear and non-linear voxel-wise heads trained on top of the frozen \texttt{vox2vec} representations (linear and non-linear probing) perform on par with supervised end-to-end FPN. From left to right: a) a slice of the input volume; predictions of b) linear voxel-wise head on top of \texttt{vox2vec}, c) non-linear voxel-wise head on top of \texttt{vox2vec}, d) supervised end-to-end FPN; e) ground truth segmentation.}
    \label{fig:example}
\end{figure}

%% file: paper3421.bbl
\begin{thebibliography}{10}
\providecommand{\url}[1]{\texttt{#1}}
\providecommand{\urlprefix}{URL }
\providecommand{\doi}[1]{https://doi.org/#1}

\bibitem{nlst}
Data from the national lung screening trial (nlst) (2013).
  \doi{10.7937/TCIA.HMQ8-J677},
  \url{https://wiki.cancerimagingarchive.net/x/-oJY}

\bibitem{transvw}
Transferable visual words: Exploiting the semantics of anatomical patterns for
  self-supervised learning. IEEE Transactions on Medical Imaging
  \textbf{40}(10),  2857--2868 (Oct 2021). \doi{10.1109/TMI.2021.3060634}

\bibitem{nsclc}
Aerts, H., Velazquez, E.R., Leijenaar, R., Parmar, C., Grossmann, P., Cavalho,
  S., Bussink, J., Monshouwer, R., Haibe-Kains, B., Rietveld, D., et~al.: Data
  from nsclc-radiomics. The cancer imaging archive  (2015)

\bibitem{msd}
Antonelli, M., Reinke, A., Bakas, S., Farahani, K., Kopp-Schneider, A.,
  Landman, B.A., Litjens, G., Menze, B., Ronneberger, O., Summers, R.M.,
  et~al.: The medical segmentation decathlon. Nature communications
  \textbf{13}(1), ~4128 (2022)

\bibitem{lidc}
Armato~III, S.G., McLennan, G., Bidaut, L., McNitt-Gray, M.F., Meyer, C.R.,
  Reeves, A.P., Zhao, B., Aberle, D.R., Henschke, C.I., Hoffman, E.A., et~al.:
  The lung image database consortium (lidc) and image database resource
  initiative (idri): a completed reference database of lung nodules on ct
  scans. Medical physics  \textbf{38}(2),  915--931 (2011)

\bibitem{vicregl}
Bardes, A., Ponce, J., LeCun, Y.: {VICRegL: Self-Supervised Learning of Local
  Visual Features}. arXiv  (Oct 2022). \doi{10.48550/arXiv.2210.01571}

\bibitem{global_and_local}
Chaitanya, K., Erdil, E., Karani, N., Konukoglu, E.: Contrastive learning of
  global and local features for medical image segmentation with limited
  annotations. Advances in Neural Information Processing Systems  \textbf{33},
  12546--12558 (2020)

\bibitem{simclr}
Chen, T., Kornblith, S., Norouzi, M., Hinton, G.: A simple framework for
  contrastive learning of visual representations. In: International conference
  on machine learning. pp. 1597--1607. PMLR (2020)

\bibitem{simsiam}
Chen, X., He, K.: Exploring simple siamese representation learning. In:
  Proceedings of the IEEE/CVF conference on computer vision and pattern
  recognition. pp. 15750--15758 (2021)

\bibitem{context_patch}
Doersch, C., Gupta, A., Efros, A.A.: Unsupervised visual representation
  learning by context prediction. In: Proceedings of the IEEE international
  conference on computer vision. pp. 1422--1430 (2015)

\bibitem{catastrophic}
French, R.M.: Catastrophic forgetting in connectionist networks. Trends in
  cognitive sciences  \textbf{3}(4),  128--135 (1999)

\bibitem{contrastive}
Hadsell, R., Chopra, S., LeCun, Y.: {Dimensionality Reduction by Learning an
  Invariant Mapping}. In: {2006 IEEE Computer Society Conference on Computer
  Vision and Pattern Recognition (CVPR'06)}, vol.~2, pp. 1735--1742. IEEE (Jun
  2006). \doi{10.1109/CVPR.2006.100}

\bibitem{mae}
He, K., Chen, X., Xie, S., Li, Y., Doll'ar, P., Girshick, R.B.: Masked
  autoencoders are scalable vision learners. 2022 IEEE/CVF Conference on
  Computer Vision and Pattern Recognition (CVPR) pp. 15979--15988 (2022)

\bibitem{nnunet}
Isensee, F., Jaeger, P.F., Kohl, S.A., Petersen, J., Maier-Hein, K.H.: nnu-net:
  a self-configuring method for deep learning-based biomedical image
  segmentation. Nature methods  \textbf{18}(2),  203--211 (2021)

\bibitem{amos}
Ji, Y., Bai, H., Yang, J., Ge, C., Zhu, Y., Zhang, R., Li, Z., Zhang, L., Ma,
  W., Wan, X., et~al.: Amos: A large-scale abdominal multi-organ benchmark for
  versatile medical image segmentation. arXiv preprint arXiv:2206.08023  (2022)

\bibitem{adam}
Kingma, D.P., Ba, J.: {Adam: A Method for Stochastic Optimization}. arXiv  (Dec
  2014). \doi{10.48550/arXiv.1412.6980}

\bibitem{rotation}
Komodakis, N., Gidaris, S.: Unsupervised representation learning by predicting
  image rotations. In: International conference on learning representations
  (ICLR) (2018)

\bibitem{kondrateva2022neglectable}
Kondrateva, E., Druzhinina, P., Dalechina, A., Shirokikh, B., Belyaev, M.,
  Kurmukov, A.: Neglectable effect of brain mri data prepreprocessing for tumor
  segmentation. arXiv preprint arXiv:2204.05278  (2022)

\bibitem{btcv}
Landman, B., Xu, Z., Igelsias, J., Styner, M., Langerak, T., Klein, A.: Miccai
  multi-atlas labeling beyond the cranial vault--workshop and challenge. In:
  Proc. MICCAI Multi-Atlas Labeling Beyond Cranial Vault—Workshop Challenge.
  vol.~5, p.~12 (2015)

\bibitem{semi1}
Lee, C.E., Chung, M., Shin, Y.G.: {Voxel-level Siamese Representation Learning
  for Abdominal Multi-Organ Segmentation}. Comput. Methods Programs Biomed.
  \textbf{213},  106547 (Jan 2022). \doi{10.1016/j.cmpb.2021.106547}

\bibitem{flare}
Ma, J., Zhang, Y., Gu, S., An, X., Wang, Z., Ge, C., Wang, C., Zhang, F., Wang,
  Y., Xu, Y., et~al.: Fast and low-gpu-memory abdomen ct organ segmentation:
  the flare challenge. Medical Image Analysis  \textbf{82},  102616 (2022)

\bibitem{jigsaw}
Noroozi, M., Favaro, P.: Unsupervised learning of visual representations by
  solving jigsaw puzzles. In: European conference on computer vision. pp.
  69--84. Springer (2016)

\bibitem{vader}
O~Pinheiro, P.O., Almahairi, A., Benmalek, R., Golemo, F., Courville, A.C.:
  Unsupervised learning of dense visual representations. Advances in Neural
  Information Processing Systems  \textbf{33},  4489--4500 (2020)

\bibitem{unet}
Ronneberger, O., Fischer, P., Brox, T.: U-net: Convolutional networks for
  biomedical image segmentation. In: Medical Image Computing and
  Computer-Assisted Intervention--MICCAI 2015: 18th International Conference,
  Munich, Germany, October 5-9, 2015, Proceedings, Part III 18. pp. 234--241.
  Springer (2015)

\bibitem{3d_ssl}
Taleb, A., Loetzsch, W., Danz, N., Severin, J., Gaertner, T., Bergner, B.,
  Lippert, C.: 3d self-supervised methods for medical imaging. Advances in
  Neural Information Processing Systems  \textbf{33},  18158--18172 (2020)

\bibitem{swin_unetr}
Tang, Y., Yang, D., Li, W., Roth, H.R., Landman, B., Xu, D., Nath, V.,
  Hatamizadeh, A.: Self-supervised pre-training of swin transformers for 3d
  medical image analysis. In: Proceedings of the IEEE/CVF Conference on
  Computer Vision and Pattern Recognition. pp. 20730--20740 (2022)

\bibitem{midrc}
Tsai, E., Simpson, S., Lungren, M.P., Hershman, M., Roshkovan, L., Colak, E.,
  Erickson, B.J., Shih, G., Stein, A., Kalpathy-Cramer, J., Shen, J., Hafez,
  M.A., John, S., Rajiah, P., Pogatchnik, B.P., Mongan, J.T., Altinmakas, E.,
  Ranschaert, E., Kitamura, F.C., Topff, L., Moy, L., Kanne, J.P., Wu, C.C.:
  Medical imaging data resource center - rsna international covid radiology
  database release 1a - chest ct covid+ (midrc-ricord-1a) (2020).
  \doi{10.7937/VTW4-X588}, \url{https://wiki.cancerimagingarchive.net/x/DoDTB}

\bibitem{propagate_yourself}
Xie, Z., Lin, Y., Zhang, Z., Cao, Y., Lin, S., Hu, H.: Propagate yourself:
  Exploring pixel-level consistency for unsupervised visual representation
  learning. In: Proceedings of the IEEE/CVF Conference on Computer Vision and
  Pattern Recognition. pp. 16684--16693 (2021)

\bibitem{sam}
Yan, K., Cai, J., Jin, D., Miao, S., Guo, D., Harrison, A.P., Tang, Y., Xiao,
  J., Lu, J., Lu, L.: Sam: Self-supervised learning of pixel-wise anatomical
  embeddings in radiological images. IEEE Transactions on Medical Imaging
  (2022)

\end{thebibliography}
